\documentclass[12pt]{article}
\usepackage{cite,graphicx,epsfig}
\renewcommand{\baselinestretch}{1.5}

\title{Beyond Feedforward Models Trained by Backpropagation: a Practical Training Tool for a More Efficient Universal Approximator}
\author{{Roman~Ilin\thanks{Roman Ilin is with Department of Computer Science at The University of Memphis, Memphis, TN 38117. E-mail: rilin@memphis.edu}}
, {Robert~Kozma\thanks{Robert Kozma is with Department of Computer
Science at The University of Memphis, Memphis, TN 38117. E-mail:
rkozma@memphis.edu}}, {Paul~J.~Werbos\thanks{Paul J. Werbos, Room
675, National Science Foundation, Arlington, VA 22230. E-mail:
pwerbos@nsf.gov}\thanks{The opinions expressed in this paper are
of the authors and  do not necessarily reflect the views of their
employers, in particular NSF}}}

\begin{document}
\maketitle

Paper to appear in IEEE Transactions on Neural Networks

\section*{Abstract}
Cellular Simultaneous Recurrent Neural Network (SRN) has been
shown to be a function approximator more powerful than the MLP.
This means that the complexity of MLP would be prohibitively large
for some problems while SRN could realize the desired mapping with
acceptable computational constraints. The speed of training of
complex recurrent networks is crucial to their successful
application. Present work improves the previous results by
training the network with extended Kalman filter (EKF). We
implemented a generic Cellular SRN and applied it for solving two
challenging problems: 2D maze navigation and a subset of the
connectedness problem.  The speed of convergence has been improved
by several orders of magnitude in comparison with the earlier
results in the case of maze navigation, and superior
generalization has been demonstrated in the case of connectedness.
The implications of this improvements are discussed.

\newpage
\section{Introduction} \label{section_Intro}
The artificial neural networks, inspired by the enormous
capabilities of living brains, are one of the cornerstones of
today's field of artificial intelligence. Their applicability to
real world engineering problems has become unquestionable in the
recent decades, see for example \cite{WhiteSofgeHandbook}. Yet
most of the networks used in the real world applications use the
feed-forward architecture, which is a far cry from the massively
recurrent architecture of the biological brains.  The widespread
use of feed-forward architecture is facilitated by the
availability of numerous efficient training methods. However, the
introduction of recurrent elements makes training more difficult
and even impractical for most non-trivial cases.

The SRN's have been shown to be more powerful function
approximators by several researchers (\cite{WerbosPang96},
\cite{MLP_SRN_GANESH}). It has been shown experimentally that an
arbitrary function generated by a MLP can always be learned by an
SRN.  However the opposite was not true, as not all functions
given by a SRN could be learned by a MLP.  These results support
the idea that the recurrent networks are essential in harnessing
the power of brain-like computing.

It is well known that MLPs and a variety of kernel-based networks
(like RBF) are universal function approximators, in some sense.
Andrew Barron \cite{ABarronApproxBounds} proved that MLPs are
better than linear basis function systems like Taylor series in
approximating smooth functions; more precisely, as the number of
inputs N to a learning system grows, the required complexity for
an MLP only grows as $O(N)$, while the complexity for a linear
basis function approximator grows exponentially, for a given
degree of accuracy in approximation. However, when the function to
be approximated does not live up to the usual concept of
smoothness, or when the number of inputs becomes even larger than
what an MLP can readily handle, it becomes ever more important to
use a more general class of neural network.

The area of intelligent control provides examples of very
difficult functions to be tackled by ANN's.  Such functions arise
as solutions to multistage optimization problems, given by the
Bellman equation \ref{eq_BellmanEquaion}.  The design of
non-linear control systems, also known as "Adaptive Critics",
presupposes the ability of the so called "Critic network" to
approximate the solution of the Bellman equation. See
\cite{ProkhACD} for overview of adaptive critic designs.  Such
problems also are classified as Approximate Dynamic Programming
(ADP).  A simple example of such function is the 2D maze
navigation problem, considered in this contribution. See
\cite{WerbosPangArXiv} for in depth overview of the ADP and Maze
navigation problem.  The applications of EKF for NN training have
been developed by researchers in the field of control
\cite{PuskFeldRNNTraining}, \cite{Feldkamp}, \cite{ProkhACD}.

The classic challenge posed by Rosenblatt to perception theory is
the recognition of topological relations
\cite{RosenblattPrinciples}.  Minsky and Papert
\cite{MinskyPapert} have shown that such problems fundamentally
cannot be solved by perceptrons because of their exponential
complexity.  The multi-layer perceptrons are more powerful than
Rosenblatt's perceptron but they are also claimed to be
fundamentally limited in their ability to solve topological
relation problems \cite{MinskyPapertExp}.  An example of such
problem is the connectedness predicate.  The task is to determine
whether the input pattern is connected regardless of its shape and
size.

The two problems described above pose fundamental challenges to
the new types of neural networks, just like the XOR problem posed
a fundamental challenge to the perceptrons, which could be
overcome only by the introduction of the hidden layer and thus
effectively moving to the new type of ANN.

In this contribution, we present the Cellular Simultaneous Neural
Network (CSRN) architecture. This is a case of more generic
architecture called ObjectNet, see
\cite{NeuroDynBookPerlovskyKozma}, chapter 6, page 120. We use the
Extended Kalman Filter (EKF) methodology for training our networks
and obtain very encouraging results. For the first time an
efficient training methodology is applied to the complex recurrent
network architecture.  Extending the preliminary result introduced
in \cite{IlinKozmaWerbosMaze}, the present study addresses not
only learning but also generalization of the network on two
problem: maze and connectedness. Improvement in speed of learning
by several orders of magnitude as a result of using EKF is also
demonstrated.   We consider the results introduced in this work as
initial demonstration of the proposed learning principle, which
should be thoroughly studied and implemented in various domains.

The rest of this paper is organized as follows. Section
\ref{section_BackProp} describes the calculation of derivatives in
the recurrent network.  Section \ref{section_CSRN} describes the
CSRN architecture.  Section \ref{section_EKF} gives the EKF
formulas.  Section  \ref{section_CRSNAlgo} describes the operation
of a generic CSRN application.  Sections
\ref{section_Applications} and \ref{section_Results} describe the
two problems addressed by this contribution and give the
simulation results. Section \ref{section_Discusion} is the
discussion  and conclusions.

\section{Backpropagation in complex networks} \label{section_BackProp}
The backpropagation algorithm is the foundation of neural network
applications \cite{WERBOS_BPTT}, \cite{WERBOS1}.  Backpropagation
relies on the ability to calculate the exact derivatives of the
network outputs with respect  to all the network parameters.

Real live applications often demand complex networks with large
number of parameters.  In such cases, the use of the rule of
ordered derivatives \cite{WERBOS_BPTT}, \cite{FeldKampProchPhased}
allows to obtain the derivatives in systematic manner. This rule
also allows to simplify the calculations by breaking a complex
network into simple building blocks, each characterized by its
inputs, outputs and parameters. If the derivatives of the outputs
of a simple building block with respect to all its internal
parameters and inputs are known, then the derivatives of the
complete system can be easily obtained by backpropagating through
each block.

Suppose that the network consists of N units, or subnetworks,
which are updated in order from 1 to N. We would like to know the
derivatives of the network outputs with respect to the parameters
of each unit. In general case the final calculation for any
network output $j$ is a simple summation:

\begin{equation}
\frac{\partial^+ z_j}{\partial \alpha} = \sum_{i=1}^{N}
\sum_{k=1}^{n} \delta_k^i \frac{\partial^+ z_k^i}{\partial \alpha}
\label{eq_SumTotalDerivativeOfNetwork}
\end{equation}

here $\alpha$ stands for any parameter, $i$ is the unit number,
$k$ is the index of the output of the current unit, $\delta_k^j$
is the derivative with respect to the input of the unit that is
connected to the $k^{th} output$ of the $i^{th}$ unit. Note that
the $k^{th}$ output of the current unit can feed into several
subsequent units and so the "delta" will be a sum of the "deltas"
obtained from each unit. Also $\delta_k^N$'s are set externally as
if the network were a part of a bigger system.  If we simply want
the derivatives of the outputs, set $\delta_k^N = 1$.  We provide
an example of this calculation in Appendix A.

Let's denote the outputs of our network as $z_i$. Ultimately we
are interested in obtaining the derivatives of these outputs
w.r.t. all the internal parameters.  This is equivalent to
calculating the Jacobian matrix of the system.  For example, if we
have two outputs and three internal parameters a,b, and c, the
Jacobian will be

%\[ \bar C =  \left(  \begin{array}{ccc}
%\frac{\partial^+ z_1}{\partial a}  & \frac{\partial^+ z_1}{\partial b} & \frac{\partial^+ z_1}{\partial c}\\
%\frac{\partial^+ z_2}{\partial a}  & \frac{\partial^+ z_2}{\partial b} & \frac{\partial^+ z_2}{\partial c}\\
%\end{array} \right) \].

\begin{equation}
\bar C =  \left(  \begin{array}{ccc}
\frac{\partial^+ z_1}{\partial a}  & \frac{\partial^+ z_1}{\partial b} & \frac{\partial^+ z_1}{\partial c}\\
\frac{\partial^+ z_2}{\partial a}  & \frac{\partial^+ z_2}{\partial b} & \frac{\partial^+ z_2}{\partial c}\\
\end{array} \right)
\end{equation}

This matrix can be used to adjust the system's parameters using
various methods such as gradient descent or Extended Kalman
Filter.

So far we considered multi-layered feed-forward networks.  The
methodology described above can be extended to recurrent networks.
Let us consider a feed-forward network with recurrent connections
that link some of its outputs to some of its inputs.  Suppose that
the network is updated for N steps.  We would like to calculate
the derivatives of the final network outputs w.r.t. the weights of
the network.  This calculation is a case of Eq.
\ref{eq_SumTotalDerivativeOfNetwork}.  Suppose that the network
has $m$ inputs, and $n$ outputs. We assume that the expressions
for the derivatives of all outputs w.r.t. each input and each
network weight $\partial z_k/\partial \alpha, k=1..n$, $\partial
z_k/\partial x_p, k=1..n, p=1..m$ are known and we will also
denote the ordered derivatives $\partial^+ z_k/\partial \alpha$ by
$F^k_\alpha$. Then the full derivatives calculation is given by
the algorithm in Fig. \ref{fig_AlgoDerivativesInRNN}. Note that we
omit the loop over all the weight parameters $\alpha$ to improve
readability. The result of this algorithm is the Jacobian matrix
of the network after N iterations.

\section{Cellular Simultaneous Recurrent Networks}\label{section_CSRN}
SRN's can be used for static functional mapping, similarly to the
MLP's. They differ from more widely known time lagged recurrent
networks (TLRN) because the input in SRN is applied over many time
steps and the output is read after the initial transitions have
disappeared and the network is in equilibrium state.  The most
critical difference between TLRN and SRN is whether the network
output is required at the same time step (TLRN) or after the
network settles to an equilibrium (SRN).

Many real live problems require to process patterns that form a 2D
grid. For instance, such problems arise in image processing or in
playing a game of chess. In those cases the structure of the
neural network should also become a 2D grid. If we make all the
elements of the grid identical, the resulting cellular neural
network benefits from greatly reduced number of independent
parameters.

\begin{figure}[htp]
\centerline{\includegraphics[width=5in]{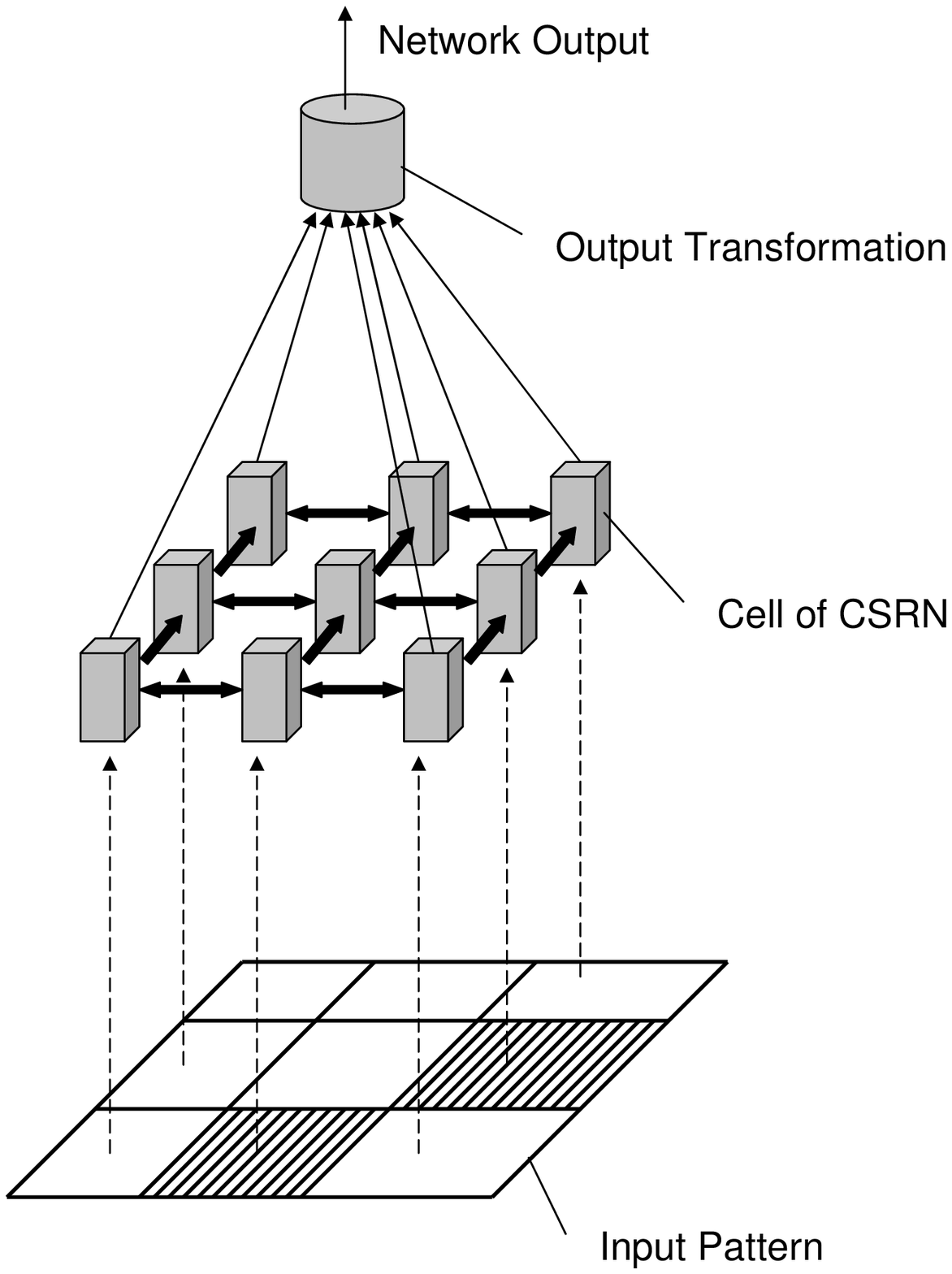}}
\caption{} \label{fig_CSRNArchitecture}
\end{figure}

The combination of cellular structure with SRN creates very
powerful function approximators.  We developed a CSRN package that
can be easily adopted to various problems.  The architecture of
the network is given in Fig. \ref{fig_CSRNArchitecture}.  The
input is always a 2D grid. The number of cells in the network
equals to the size of the input. The number of outputs also equals
to the size of the input.  Since most of the problems require only
few outputs, we add an arbitrary output transformation to the
network. It has to be differentiable, but it does not have to have
adjustable parameters. The training in our applications occurs
only in the SRN.  The cells of the network are connected through
neighbor links. Each cell has four neighbors and the edges of the
network wrap around.

\begin{figure}[htp]
\centerline{\includegraphics[width=5in]{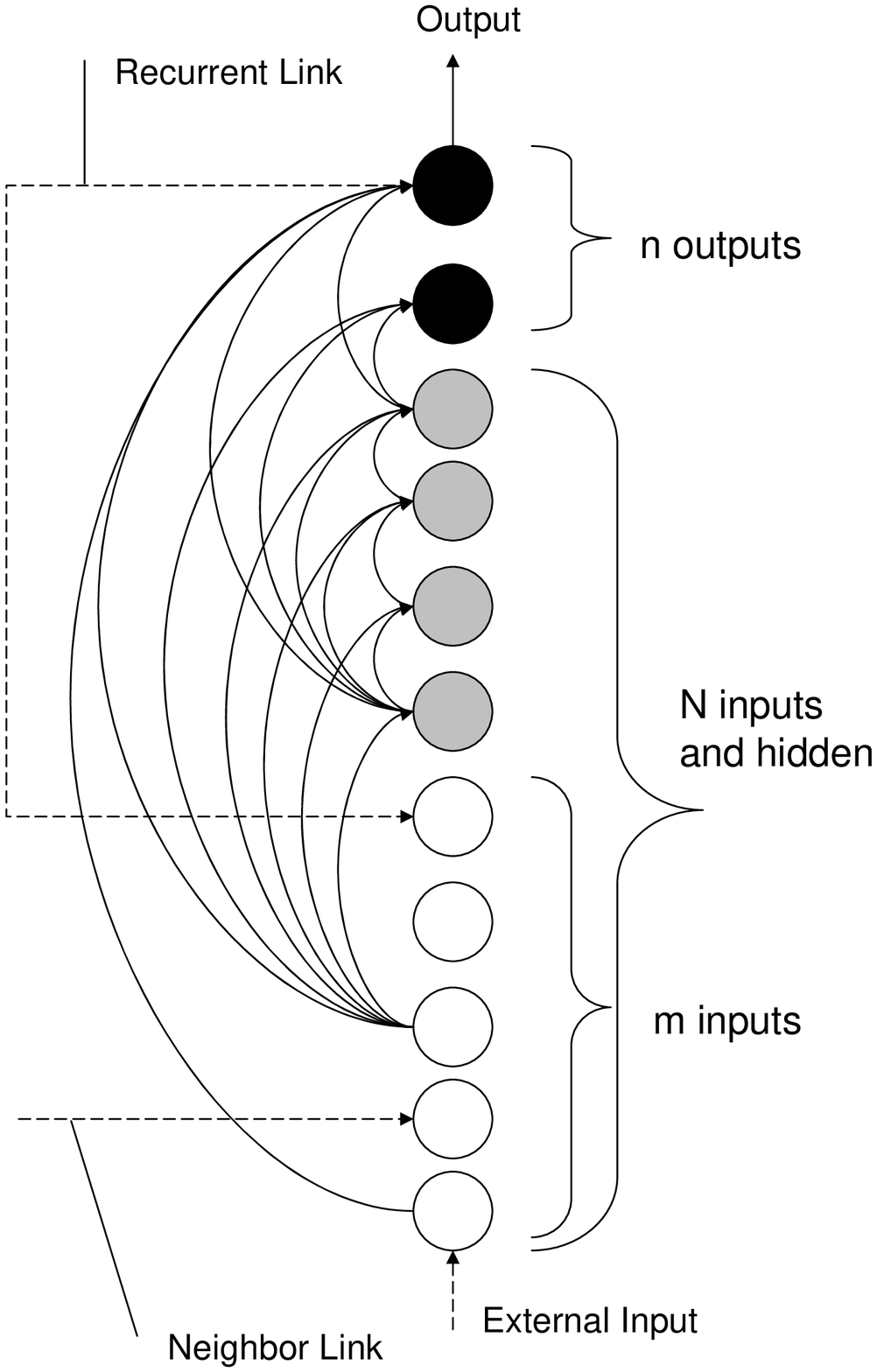}} \caption{}
\label{fig_CellGMLP}
\end{figure}

The cell of CSRN in our implementation is a generalized MLP
\cite{WhiteSofgeCh3}, shown in Fig. \ref{fig_CellGMLP}.  Each
non-input node of the GMLP is linked to all the subsequent nodes
thus generalizing the idea of multi-layered network.  The
recurrent connections come from the output nodes of this cell and
from its neighboring cells.  It is an important feature of the
used architecture that each cell has the same weights, which
allows to build arbitrary large networks without increasing the
number of weight parameters.

\section{Extended Kalman Filter for Network Training}\label{section_EKF}
Kalman filters (KF) originated in signal processing.  They present
a computational  technique which allows to estimate the hidden
state of a system based on observable measurements.  See
\cite{MonashUnivKalman}, \cite{Anderson} for derivation of the
classical Kalman filter formulas based on the theory of
multivariate normal distribution.

In the case of neural network training, we are faced with the
problem of determining the parameter weights in such a way that
the measured outputs of the network are as close to the target
values as possible. The network can be described as a dynamical
system with its hidden state vector $\vec W$ formed by all the
values of network weights, and the observable measurements vector
formed by the values of network outputs $\vec Y$.  It is sometimes
convenient to form a full state vector $\vec S$ that consists of
both hidden and observable parts.  Such formulation can be used in
the derivation of Kalman filter \cite{MonashUnivKalman}.  In this
paper we follow the convention in the literature which refer to
$\vec W$ as the state vector \cite{SHaykinKalmanFilters}. Note
that the outputs of the network can be expressed in terms of the
weights as

%\begin{equation}
%\vec S = \left(  \begin{array}{c}
%\vec W\\
%\vec Y\\
%\end{array} \right)
%\label{eq_CSRNStatevector}
%\end{equation}

\begin{equation}
 \vec Y = \bar C \vec W
\end{equation}

where $\bar C$ is the Jacobian matrix of the network evaluated
around the current output vector $\vec Y$. We assume that the
state $\vec S$ is normally distributed and we are interested in
the estimate of $\vec W$ based on the knowledge of $\vec Y$ and
the underlying dynamical model of the system, which is simply
$\vec Y(i+1) =\vec t$ and  $\vec W(i+1) = \vec W(i)$, where $\vec
t$ is the target output of the network. Suppose the covariance
matrix of $\vec W$ is given by $\bar K$, and the measurement noise
covariance matrix by $\bar R$.  $\bar R$ is assumed to be normally
distributed with zero mean. Then the Kalman update is given by the
following formulas.

\begin{equation}
\vec W(i+1) = \vec W(i) + \frac {\bar K(i) \bar C(i)^T} {\bar C(i)
\bar K(i) \bar C(i)^T + \bar R(i)} (\vec t - \vec Y(i) )
\label{eq_EKFMean}
\end{equation}

\begin{equation}
\bar K(i+1) = \bar K(i) - \frac {\bar K(i) \bar C(i)^T \bar C(i)
\bar K(i)} {\bar C(i) \bar K(i) \bar C(i)^T + \bar R(i)} + \bar
Q(i) \label{eq_EKFCovariance}
\end{equation}

We introduced the index i to denote current training step. The
matrix $\bar Q(i)$ is the process noise covariance matrix. It
represents our assumptions about the distribution of the true
values of $\vec W$.  The formulas \ref{eq_EKFMean} and
\ref{eq_EKFCovariance} are the celebrated Extended Kalman filter
formulas which can be found in numerous literature, see for
example \cite{SHaykin}, \cite{SHaykinKalmanFilters} ,
\cite{Feldkamp}.  If we look closely at \ref{eq_EKFMean}, we can
see the similarity between the EKF update and the regular gradient
descend update. In both cases we have some matrix coefficient
multiplied by the difference $(\vec t - \vec Y(i))$.  In the case
of gradient descend , the coefficient is simply $\bar C(i)$
multiplied by some learning rate.  In the case of EKF, the
coefficient is more complex as it involves the covariance matrix
$\bar K$, which is the key to the efficiency of EKF.

The process noise $\bar Q$ can be safely assumed to be 0, even
though setting it to a non-zero value helps prevent $\bar K$ from
becoming negative definite and destabilizing the filter. The
measurement noise $\bar R$ also plays an important role in fine
tuning the EKF by accelerating the speed of learning. The proper
functioning of EKF depends on the assumption that the state vector
$\vec S$ is normally distributed. This assumption usually does not
hold in practice. However, adding the normally distributed noise
described by $\bar R$ helps overcome this difficulty.  $\vec R$ is
usually chosen to be a random diagonal matrix.  The values on the
diagonal are annealed as the network training progresses, so that
by the end of training, noise is insignificant.  It turned out
that the way $R$ is annealed has significant effect on the rate of
convergence. After experimenting with different functional forms
we used the following formula:

\begin{equation}
R(i) = alog(b\vec \delta(i)^2 +1) I \label{eq_Rannealing}
\end{equation}

where $\delta(i)^2$ is the squared error, $\delta(i) =\vec t -
\vec Y(i)$.  The constants $a$ and $b$ were determined
experimentally. We used $a=b=0.001$, which gave reasonably good
results presented in the next sections. This functional form works
better than linear annealing. Making the measurement noise a
function of the error results in fast and reliable learning.

The algorithm described above is suitable for learning one
pattern. Learning multiple patterns creates additional challenges.
The patterns can be learned one by one or in a batch. In the
present work we used the batch mode which results in efficient
learning but is computationally demanding.  To explain this
method, we write Eq. \ref{eq_EKFMean} more compactly as

\begin{equation}
\vec {\delta W} = \bar G \vec \delta
\end{equation}

where $\bar G = \bar K \bar C^T / (\bar C \bar K \bar C^T + \bar
R)$  and the time step index is omitted for clarity. The matrix
$\bar G$ is called Kalman gain.

Suppose that the network has s outputs and p weights.  The size of
matrix $\bar C$ is s by p, and the size of $\bar G$ is p by s.
Suppose we have M patterns in a batch. If the network is
duplicated M times, the resulting network will have $M \times s$
outputs. The size of $\bar C$ becomes $M \times s$ by p.  Note
that we simply concatenate M matrices together. Matrix $\bar G$
can still be computed from Eq. \ref{eq_EKFMean} and its size
becomes p by $M \times s$. The weight update can be done just like
in the case of one pattern, except now the matrix $\bar G$ encodes
information about all patterns.

This method is called multi-streaming \cite{HaykinKF},
\cite{ProckMultistream}.  Increasing number of input patterns will
result in large sizes of $\bar C$ and $\bar G$.  This will make
batch update inefficient because of the need to invert large
matrices. Therefore, larger problems will require more advanced
numerical techniques already used by practitioners of EKF training
\cite{PuskFeldAvoidMatInversions}, \cite{HaykinKF}.

\section{CSRN training algorithm }\label{section_CRSNAlgo}
The network architecture given in Fig. \ref{fig_CSRNArchitecture}
is very generic and with proper implementation can be easily
adopted to different problems.   The algorithm given in Fig.
\ref{fig_CSRNAlgorithm} describes the training of CSRN.   The main
loop calculates the Jacobian matrix $\bar C$, which is used in the
Kalman weight update.  We perform testing during each training
period and make a decision to stop training based on testing
results. Output transformation is the part of the network that has
to be customized for each problem.  Other network parameters
include network size, cell size, number of internal steps of SRN,
EKF parameters $\bar K$, $\bar R$, and $\bar Q$. The number of
internal steps is selected large enough to allow the typical
network to settle to an equilibrium state.  In the process of
training the network dynamics changes and sometimes it no longer
settles.  Currently we do not terminate training even if
equilibrium is not reached as such networks still achieve good
levels of generalization. The Matlab implementation is available
from the authors.

\section{Application of EKF Learning to Generalized Maze Navigation Problem}\label{section_Applications}
\subsection{Problem Description} 
\begin{figure}[htp]
\centerline{\includegraphics[width=5in]{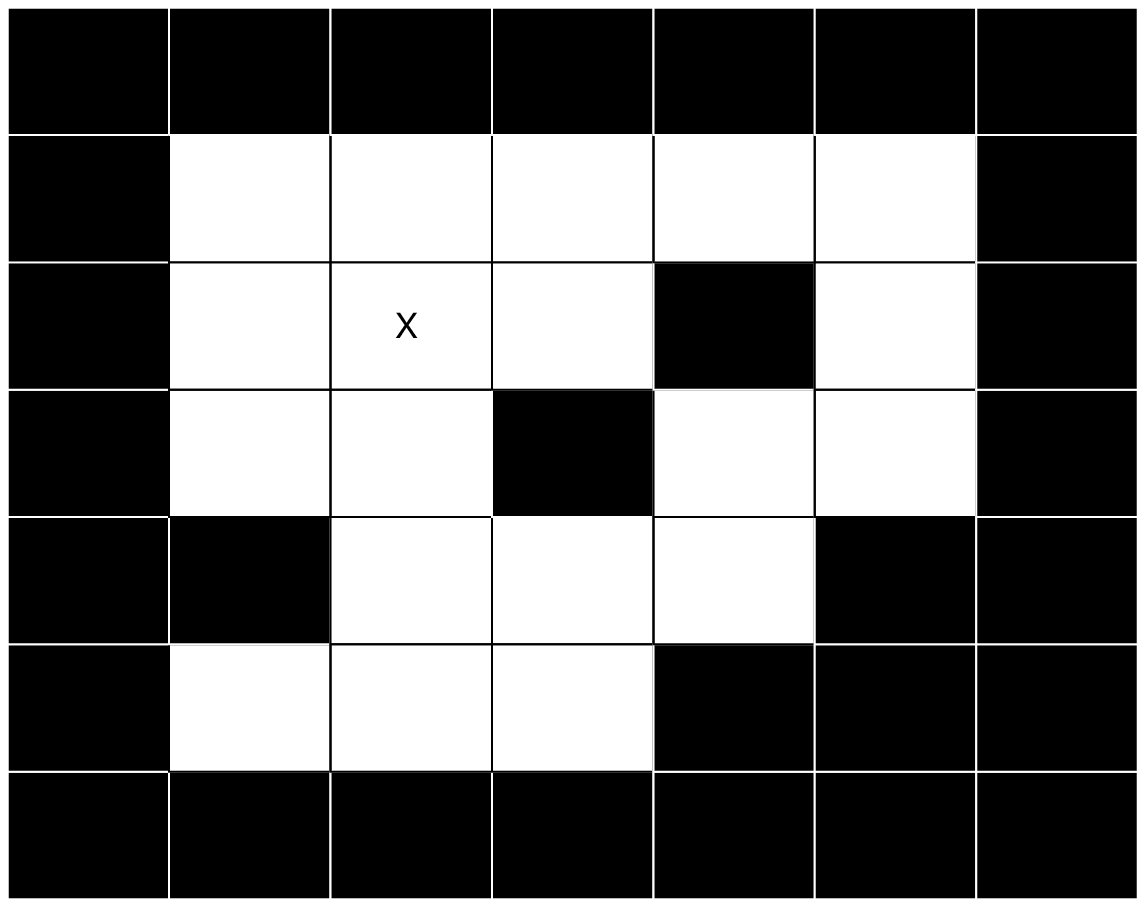}}
\caption{} \label{fig_MazeWorld}
\end{figure}

The generalized
maze navigation consists of finding the optimal path from any
initial position to the goal in a 2D grid world.  An example of
such a world is illustrated in Fig. \ref{fig_MazeWorld}.  One
version of an algorithm for solving this problem will take a
representation of the maze as its input and return the length of
path from each clear cell to the goal.  For example, for a 5 by 5
maze the output will consist of 25 numbers.  Once we know the
numbers it is very easy to find the optimal path from any cell by
simply following the minimum among the neighbors.  Examples of
such outputs are given in Fig. \ref{fig_MazeWithArrows}.

2D Maze Navigation is a very simple representative of a broad
class of problems solved using the techniques of Dynamic
Programming, which means finding the J cost-to-go function using
the Bellman's equation (see for example \cite{SHaykin}).  Dynamic
Programming gives the exact solution to multistage decision
problems.  More precisely, given a Markovian decision process with
N possible states and the immediate expected cost of transition
between any two states $i$ and $j$ denoted by $c(i,j)$, the
optimal cost-to-go function for each state satisfies the following
Bellman's optimality equation.

\begin{equation}
J^*(i) = min_{\mu}(c(i,\mu(i)) + \gamma \sum_{j=1}^{N}p_{ij}(\mu)
J^*(j)) \label{eq_BellmanEquaion}
\end{equation}

$J(i)$ is the total expected cost from the initial state $i$, and
$\gamma$ is the discount factor.  The cost $J$ depends on the
policy $\mu$, which is the mapping between the states and actions
causing state transitions.  The optimal expected cost results from
the optimal policy $\mu*$. Finding such policy directly from Eq.
\ref{eq_BellmanEquaion} is possible using recursive techniques but
computationally expensive as the number of states of the problem
grows.  In the case of 2D maze, the immediate cost $c(i,j)$ is always 1,
and the probabilities $p_{ij}$ can only take values of 0 or 1.

The J surface resulting from the 2D maze is a challenging function
to be approximated by a neural network.  It has been shown  that
an MLP cannot solve the generalized problem \cite{WerbosPang96}.
Therefore, this is a great problem to demonstrate the power of the
Cellular SRN's.   It has been shown that CSRN is capable of
solving this problem by designing its weights in a certain way
\cite{Wunsch2000}.  However the challenge is to train the network
to do the same.

The Cellular SRN to solve the m by m maze problem consists of m+2
by m+2 grid of identical units. The extra row and column on each
side result from introducing the walls around the maze which
prevent the agent from running away.  Each unit receives input
from the corresponding cell of the maze and returns the value of
the J function for this cell.  There are two inputs for each cell,
one indicates whether this is a clear cell or an obstacle and the
other supplies the values of the goal.  As shown in Fig.
\ref{fig_CSRNArchitecture}, the number of outputs of the cellular
part of the network equals to the number of cells.   In the maze
application the final output is the values of J function for each
input cell and therefore there is no need for the output
transformation.

\subsection{Results of 2D Maze Navigation}

Previous results of training the Cellular SRN's showed slow
convergence \cite{WerbosPang96}.  Those experiments used
back-propagation with adaptive learning rate (ALR)
\cite{WhiteSofgeHandbook}.  The network consisted of 5 recurrent
nodes in each cell and was trained on up to 6 mazes.  The initial
results demonstrated the ability of the network to learn the mazes
\cite{IlinKozmaWerbosMaze}.

The introduction of EKF significantly sped up the training of the
Cellular SRN. In the case of single maze, the network reliably
converges within 10-20 training cycles (see fig.
\ref{fig_CSRNAlgorithm}).  In comparison, back-propagation through
time with adaptive learning rate (ALR) takes between 500 and 1000
training cycles and is more dependent on the initial network
weights \cite{WerbosPang96}.

\begin{figure}[htp]
    \centerline{\hbox{ \hspace{0.0in}
        \epsfxsize=5in
        \epsffile{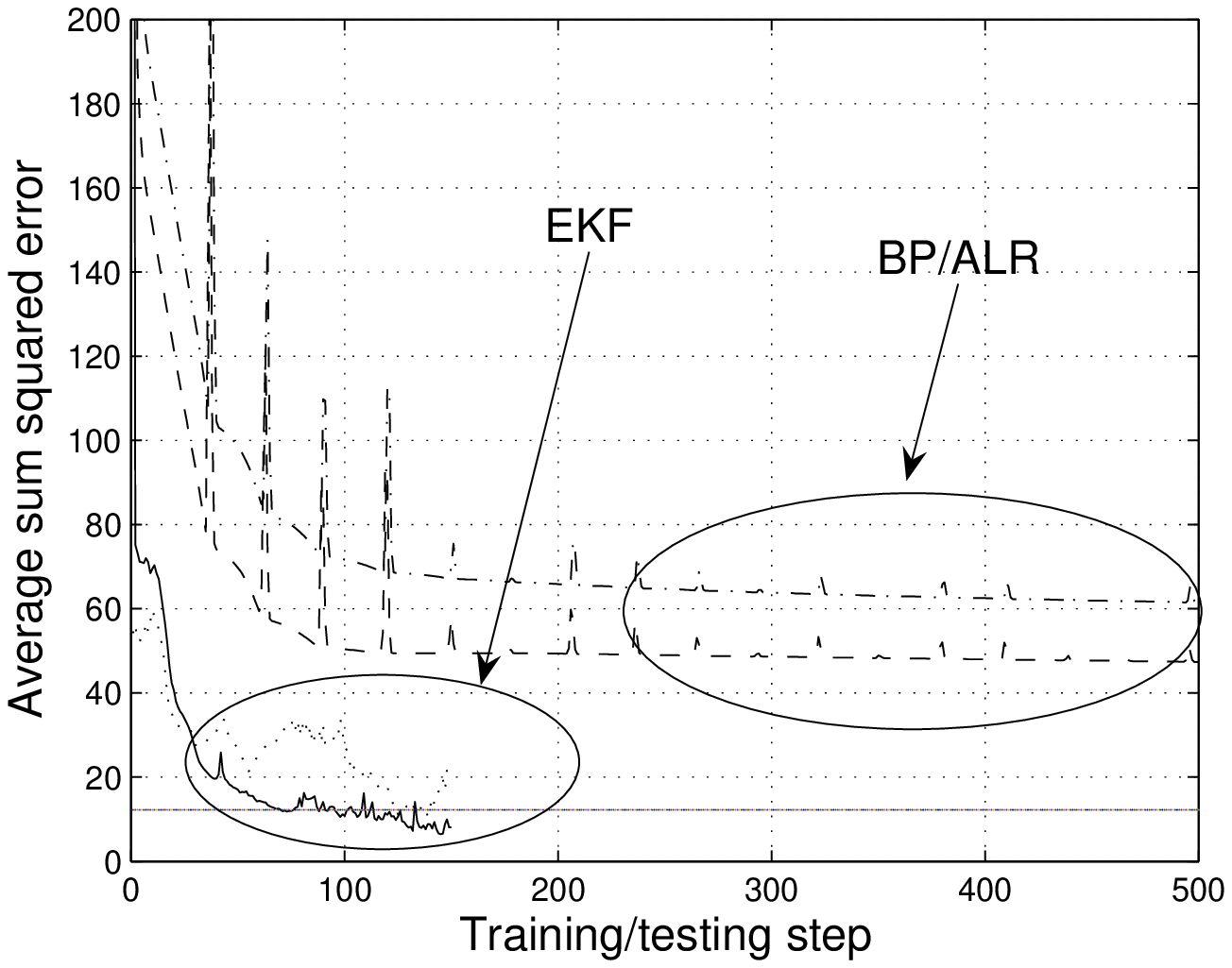}
            }
        }
    \vspace{9pt}
    \hbox{\hspace{3in} A }
    \vspace{9pt}

    \centerline{\hbox{ \hspace{0.0in}
        \epsfxsize=5in
        \epsffile{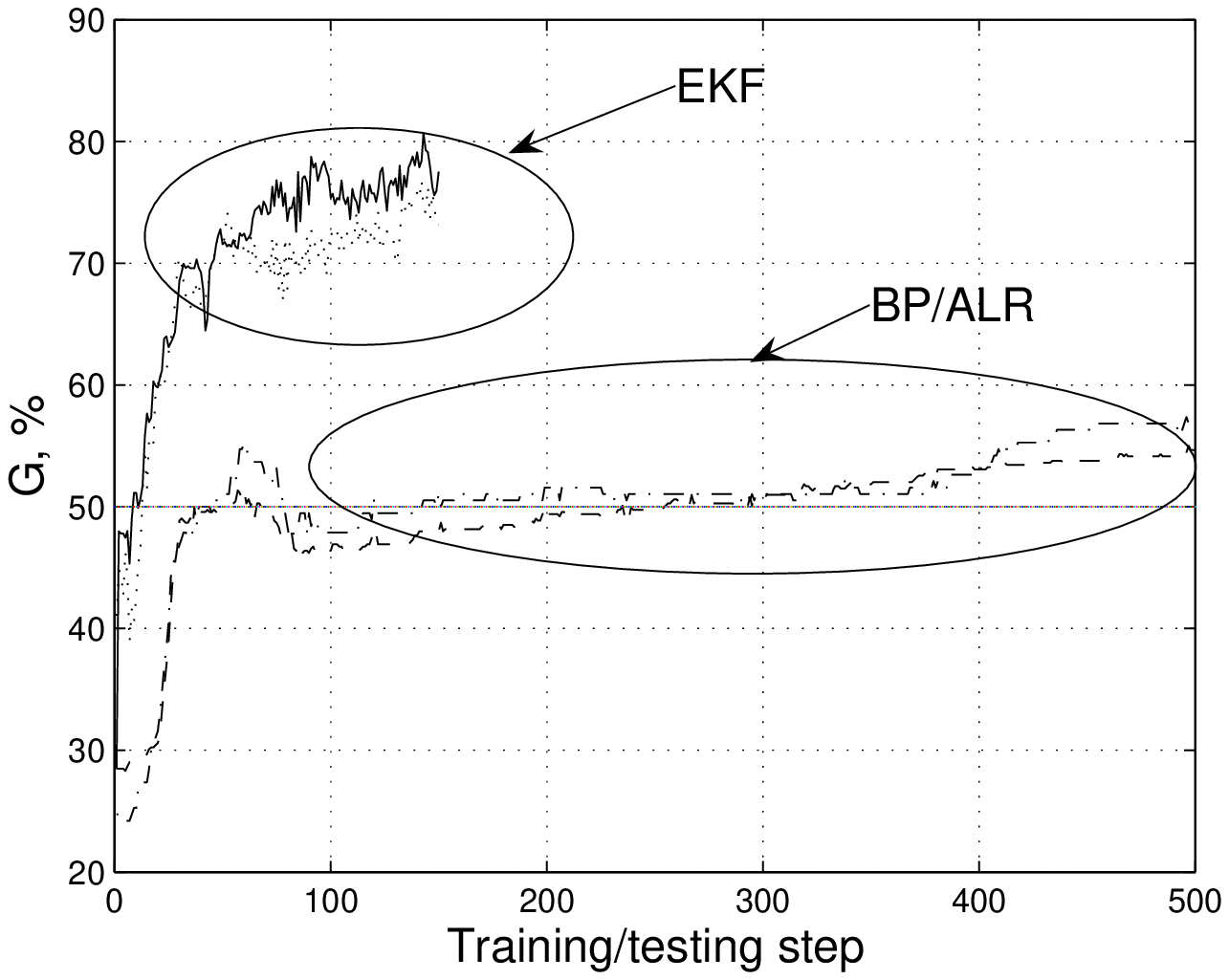}
            }
        }

    \vspace{9pt}
    \hbox{\hspace{3in} B }
    \vspace{9pt}

\caption{} \label{fig_GoodExampleAB}
\end{figure}

We discovered that increasing the number of recurrent nodes from 5
to 15 allows to speed up both EKF and ALR training in case of
multiple mazes.  Nevertheless the EKF has a clear advantage. For
more realistic learning assignment we use 30 training mazes and
test the network with 10 previously unseen mazes. The training
targets where computed using dynamic programming algorithm. Fig.
\ref{fig_GoodExampleAB}A shows the sum squared error as function
of the training step.  We can see that EKF reaches the reasonable
level in 150 training cycles.  For comparison the back-propagation
through time with adaptive learning rate training is shown on the
same graph.

\begin{figure}[htp]
\centerline{\includegraphics[width=5in]{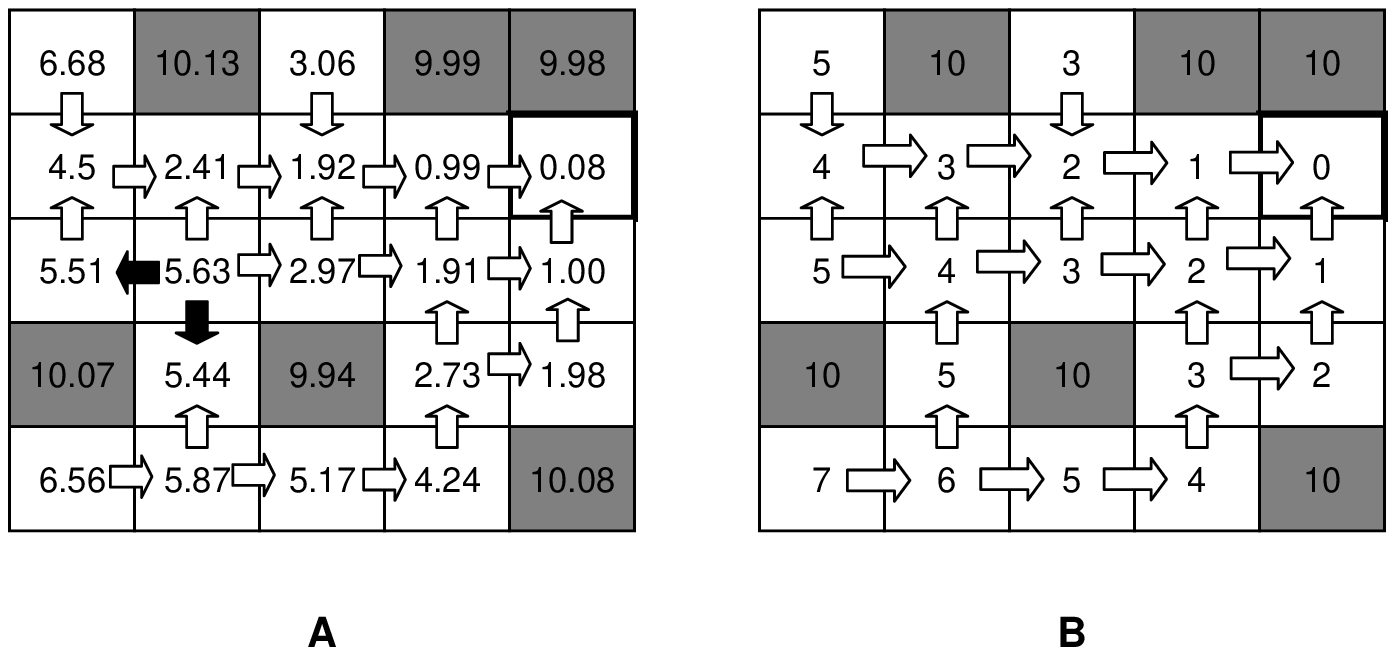}}
\caption{} \label{fig_MazeWithArrows}
\end{figure}

The true solution consists of integer values with the difference
of one between the neighboring cells. We suppose that the
approximation is reasonable if the maximum error per cell is less
than 0.5, since in this case the correct differences will be
preserved. This means that for a 7 by 7 network corresponding to 5
by 5 maze, the sum squared error has to fall below $49*0.5^2 =
12.25$.  In Fig. \ref{fig_GoodExampleAB}A the EKF drops below the
12.25 level within 150 steps while ALR testing saturates at the
level close to 50.   We use 20 internal steps within each training
cycle. In practical training scenarios the error is obviously not
the same for each cell. Detailed statistical analysis can reveal
the true nature of the expected distributions. In the present
exploratory study we don't go into the details of statistics.
Rather we introduce an empirical measure of the goodness of learnt
navigation task in the following way.   We count how many
gradients point in the correct direction. The ratio of the number
of correct gradients to the total number of gradients is our
goodness ratio $G$ that can vary from 0 to 100 percent. The
gradient of the J function gives the direction of the next move.
As an example, Fig. \ref{fig_MazeWithArrows} shows the J function
computed by a network and the true J function.  Fig.
\ref{fig_MazeWithArrows} demonstrates 2 erroneous gradient
directions. The goodness G is illustrated in Fig.
\ref{fig_GoodExampleAB}B. We can see that EKF reaches testing
performance of 75-80 percent after 150 training cycles averaged
over 10 testing mazes.  On the other hand BP/ALR testing
performance lingers around 50 percent chance level for several
hundred training cycles.  Even after 500 training cycles it is
close to the chance level. This shows the potential of EKF for
training CSRN's.

\section{Application of EKF to Connectedness Problem}\label{section_Results}
\subsection{A Simple Connectedness Problem}
\begin{figure}[htp]
\centerline{\includegraphics[width=5in]{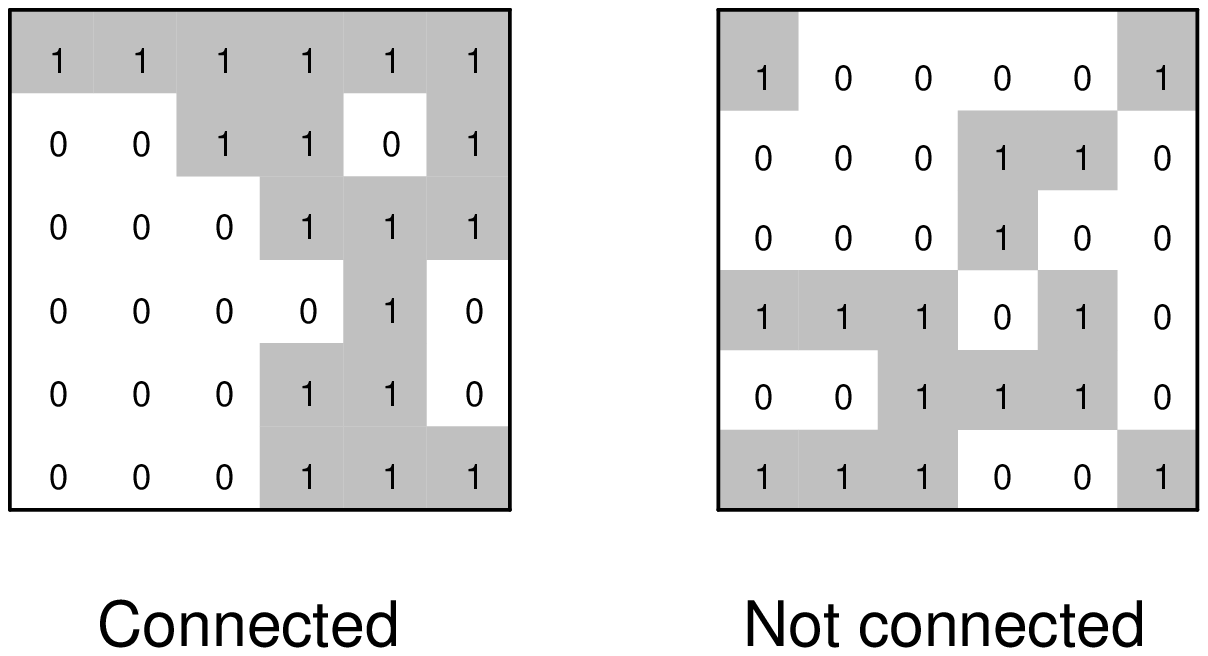}}
\caption{} \label{fig_ConnectednessExamples}
\end{figure}

The description of connectedness problem can be found in
\cite{MinskyPapert}.  The problem consists of answering the
question "is the input pattern connected?".  Such question is
fundamental to our ability to segment visual images into separate
objects which is the first preprocessing step before trying to
recognize and classify the objects.  We work with a subset of the
connectedness problem, where we consider a square image and ask
the following question: "are the top left and the bottom right
corners connected?".  Note that the diagonal connections do not
count in our case, each pixel of a connected pattern has to have a
neighbor on the left, right, top or bottom.  Examples of such
images are given in Fig. \ref{fig_ConnectednessExamples}. This
subset is still a difficult problem which could not be solved by
the feedforward network \cite{ConnectednessParodi}.  The reason
why connectedness is a difficult problem lies it its sequential
nature.  Human eye has to follow the borders of an image
sequentially in order to classify it.  This explains the need for
recursion.

The network architecture for the connectedness problem is that of
Fig. \ref{fig_CSRNArchitecture}.  The output transformation is a
GMLP with one output.  The weights of this GMLP are randomly
generated and fixed.  The target outputs are +0.5 for connected
pattern and -0.5 for disconnected pattern.

\subsection{Results of Connectedness Problem}
Here we present the results of solving the subset of connectedness
problem.  We applied the network to image sizes 5, 6, and 7.  In
each case we generated sets of 30 random connected and 30
disconnected patterns for training, and 10 connected and 10
disconnected patterns for testing. We used 20 internal iterations
and the training took between 100 and 200 training cycles. We used
the same EKF parameters as in the case of maze navigation.  After
training on 30 patterns we tested the network and calculated the
percent of correctly classified patterns.  We applied the same set
of patterns to a feed-forward network with one hidden layer. The
size of the hidden layer was varied to obtain the best results.
The results are summarized in the following table, where each
number is averaged over 10 experiments and the standard deviation
is also given.

\begin{tabular}{ccc}
\vspace{5mm}\\
\multicolumn{3}{c}{}  Table I.  Generalization with EKF learning for Connectedness problem \\
Input Size & Correctly Classified with CSRN & Correctly Classified
with MLP\\
\hline \\
5x5 & 80 $\pm$ 6 \% & 66 $\pm$ 10 \%\\
6x6 & 82 $\pm$ 6 \%& 65 $\pm$ 12 \%\\
7x7 & 88.5 $\pm$ 6 \%& 63 $\pm$ 12 \%\\
\vspace{5mm}\\
\end{tabular}

We can see that the performance of MLP is just slightly above
chance level whereas the CSRN produces correct answers in 80-90\%
of test cases on previously unseen patterns.  This performance is
likely to improve by fine tuning network parameters.

\section{Discussion and Conclusions}\label{section_Discusion}
In this contribution we presented the advantages of using the
Cellular SRN's as a more general topology compared to conventional
MLP.  We extended the previous results
\cite{IlinKozmaWerbosADPSymposium} by applying the CSRN to the
connectedness problem.   We applied an efficient learning
methodology - EKF - to the SRN and obtained very encouraging
results in terms of speed of convergence. The unit of our network
is a generalized MLP.  It can be easily substituted by any other
feed-forward computation suitable for problem at hand, without any
changes to the Cellular SRN. Now it becomes practical to use the
proposed combination of architecture and the training method to
any data that has 2D grid structure.  The network size does not
grow exponentially with the input size because of the weight
sharing.  The 100 by 100 input pattern could potentially be
processed by the CSRN with 15 units in each cell.  However large
networks still involve massive computations, which can be possibly
addressed by efficient hardware implementations \cite{ChuaCSRN}.

One example of such application is image processing. Detecting
connectedness is a fundamental challenge in this field.  We
applied our CSRN to a subset of connectedness problem with minimal
changes to the code. The results showed that CRSN is much better
at recognizing connectedness compared to feed-forward
architecture.

Another example of such data is the board games. The games of
chess and checkers have long been used as testing problems for AI.
Recently, neural networks coupled with evolutionary training
methods have been successfully applied to the checkers
\cite{FogelCheckers} and chess \cite{FogelChess}. The neural
network architecture used in those works is the case of Object Net
\cite{NeuroDynBookPerlovskyKozma}, mentioned in the introduction.
The input pattern (the chess board) is divided into spacial
components and the network is built with separate sub-units
receiving input from their corresponding components. The
interconnections between the sub-units of the network encode the
spacial relationships between different parts of the board.  The
outputs of the Object Net feed into another multilayered network
using to evaluate the overall "fitness" of the current situation
on the board.

From the above description we can see that the CSRN network is a
simplified case of the Object Net.  The Chess Object Net belongs
to the same class of multistage optimization problems, even though
it does not presently use recurrent units. The biggest difference
however, is the training method. The evolutionary computation has
proven to be able to solve the problem, however at high
computational cost.  The architecture used in this contribution
provides an efficient training method for the Object Net type of
networks with more biologically plausible training using local
derivatives information. The improved efficiency allows the use of
SRN's, which are proven to be more powerful in function
approximation than the MLP's.  Therefore, the Cellular SRN/EKF can
be applicable to many interesting problems.

This contribution builds upon several concepts: Recurrent NN's,
Simultaneous Recurrent Networks, Cellular NN's, Dynamic
Programming, Kalman Filters. After reviewing each of the concepts,
we presented a versatile recurrent neural network architecture
capable of efficient training based on EKF methodology.  We
demonstrate this novel application of EKF on the examples of the
maze navigation problem and connectedness problem.  Detailed study
of properties of EKF for CSRN training is in progress.

\section{Acknowledgements}
The authors would like to thank the anonymous reviewers for
detailed comments and useful suggestions.

\bibliography{IEEEabrv,mabib}

\renewcommand{\theequation}{A-\arabic{equation}}
% redefine the command that creates the equation no.
\setcounter{equation}{0}  % reset counter
\section*{Appendix A: Example of Calculating Ordered Derivatives}  % use *-form to suppress numbering

\begin{figure}[htp]
    \centerline{\hbox{ \hspace{0.0in}
        \epsfxsize=4in
        \epsffile{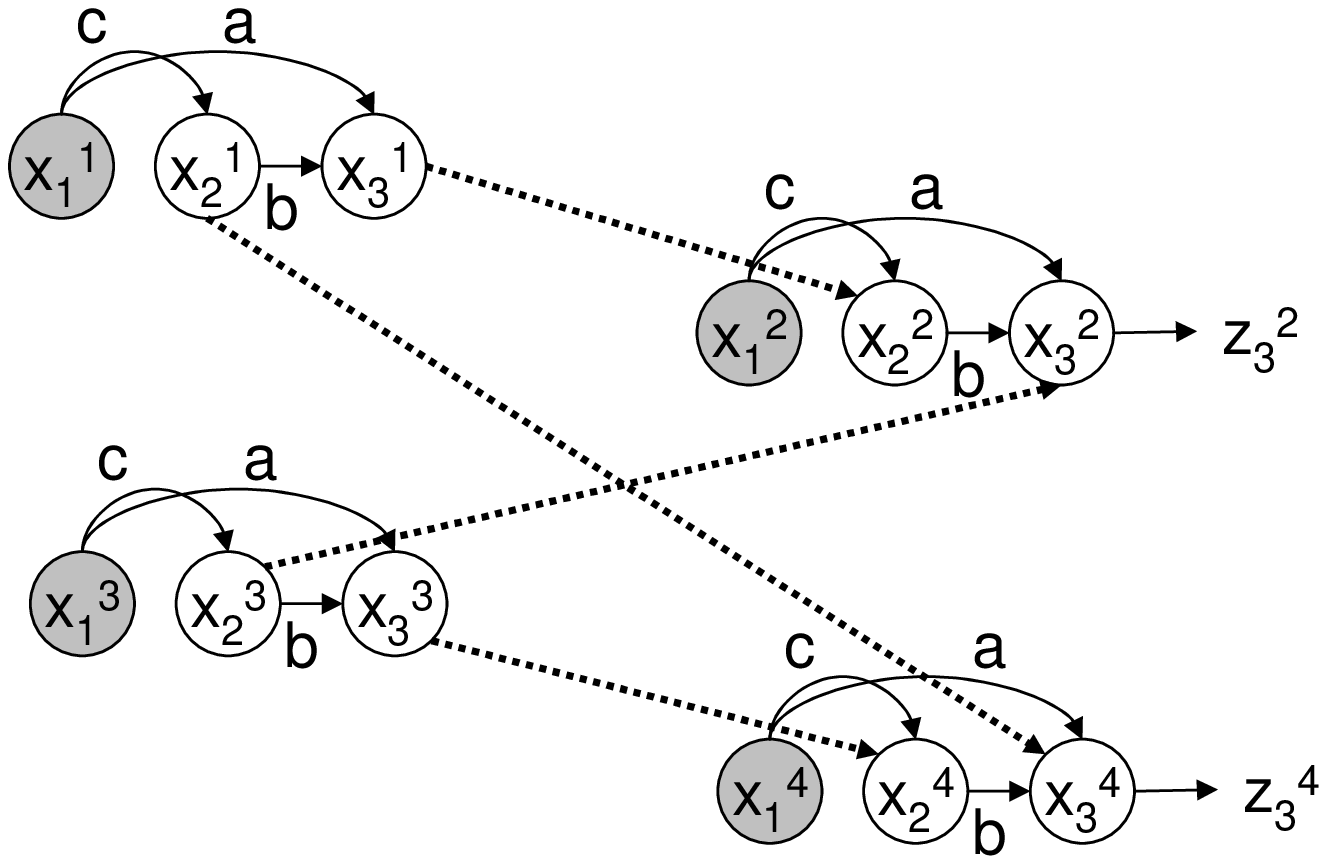}
            }
        }
    \vspace{9pt}
    \hbox{\hspace{3in} A }
    \vspace{9pt}

    \centerline{\hbox{ \hspace{0.0in}
        \epsfxsize=4in
        \epsffile{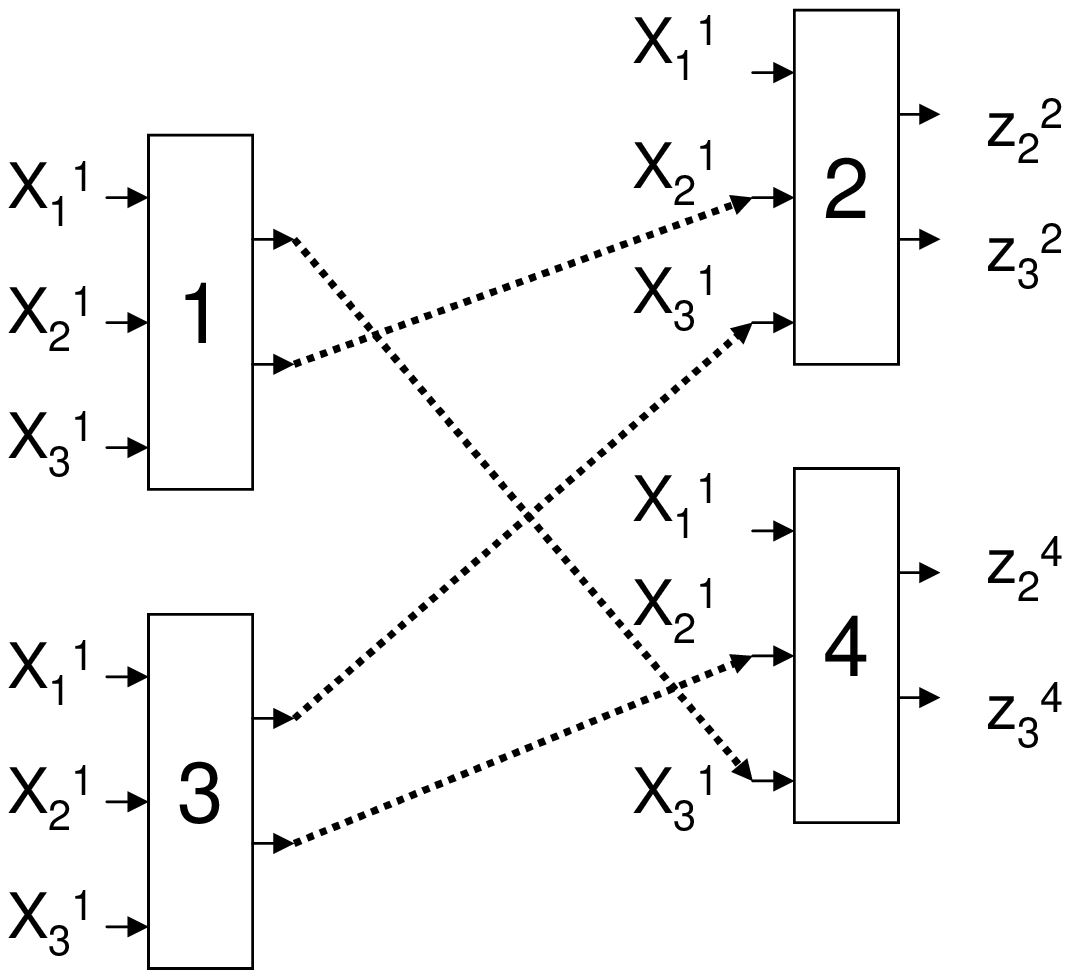}
            }
        }

    \vspace{9pt}
    \hbox{\hspace{3in} B }
    \vspace{9pt}

\caption{} \label{fig_ExampleForAppendixA}
\end{figure}

%\appendix{Appendix A: Example of Calculating Ordered Derivatives}
The following example is an illustration of the principles
mentioned in section \ref{section_BackProp}. Consider the network
in Fig. \ref{fig_ExampleForAppendixA} A. This network can be
decomposed into 4 identical units as shown in Fig.
\ref{fig_ExampleForAppendixA} B, where each unit is a mapping
between its inputs, internal parameters and outputs.  This is a
case of a simple recurrent cellular network with two cells and two
iterations.  We unfold the recurrent steps to demonstrate the
application of the rule of ordered derivatives.

Each unit has 3 inputs $x_1$,$x_3$, and $x_3$, and 3 parameters
$a$, $b$, and $c$.  The outputs of each neuron are denoted by
$z_1$, $z_2$, and $z_3$. The first input neuron does not perform
any transformation, so $z_1=x_1$.
 The second and third neurons use a nonlinear transformation $f$.
 The forward calculation of an elementary unit is as follows.

\begin{eqnarray}
z_2 &=& x_2 + f(cx_1) \\
z_3 &=& x_3 + f(ax_1 + bz_2)
 \label{eq_ForwardCell}
\end{eqnarray}

The order in which different quantities appear in the forward
calculation is
\begin{equation}
x_1, x_2, x_3, c, z_2, a, b, z_3. \label{eq_OrderOfCalc}
\end{equation}
We would like to determine the derivatives of $z_2$ and $z_3$
w.r.t. the inputs and parameters.  To do so, we apply the rule of
ordered derivatives \cite{WERBOS_BPTT} given by the following
formula.

\begin{equation}
\frac{\partial^+ TARGET}{\partial z_i} = \frac{\partial
TARGET}{\partial z_i} + \sum_{j=i+1}^{N}\frac{\partial^+
TARGET}{\partial z_j} \frac{\partial z_j}{\partial z_i}
 \label{eq_RULE_ORDRD_DER}
\end{equation}
where $TARGET$ is the variable the derivative of which w.r.t.
$z_i$ is sought, and the calculation of $TARGET$ involves using
$z_j$'s is order of their subscripts.  The notation $\partial^+$
is used for the ordered derivative, which simply means the full
derivative, as opposed to simple partial derivative obtained by
considering only the final equation involving $TARGET$.

In order to calculate the derivatives in our example, we have to
use the equation \ref{eq_RULE_ORDRD_DER} in reverse order of
\ref{eq_OrderOfCalc}.  Let $\phi$ denote the derivative of $f$.

\begin{eqnarray}
\frac{\partial^+ z_3}{\partial b} &=& \frac{\partial z_3}{\partial b} = z_2\phi(ax_1 + bz_2)\\
\frac{\partial^+ z_3}{\partial a} &=& \frac{\partial z_3}{\partial a} = x_1\phi(ax_1 + bz_2)\\
\frac{\partial^+ z_3}{\partial z_2} &=& \frac{\partial z_3}{\partial z_2} = b\phi(ax_1+bz_2)\\
\frac{\partial^+ z_3}{\partial c} &=& \frac{\partial^+ z_3}{\partial z_2}\frac{\partial^+ z_2}{\partial c} = b\phi(ax_1+bz_2)x_1\phi(cx_1)\\
\frac{\partial^+ z_3}{\partial x_3} &=& \frac{\partial z_3}{\partial x_3} = 1\\
\frac{\partial^+ z_3}{\partial x_2} &=& \frac{\partial^+
z_3}{\partial z_2} \frac{\partial z_2}{\partial x_2}
= b\phi(ax_1+bz_2)\\
\frac{\partial^+ z_3}{\partial x_1} &=& \frac{\partial^+
z_3}{\partial z_2} \frac{\partial z_2}{\partial x_1} =
b\phi(ax_1+bz_2)c\phi(cx_1)\\
\frac{\partial^+ z_2}{\partial c} &=& \frac{\partial z_2}{\partial c} = x_1\phi(cx_1)\\
\frac{\partial^+ z_2}{\partial x_3} &=& \frac{\partial z_2}{\partial x_3} = 0\\
\frac{\partial^+ z_2}{\partial x_2} &=& \frac{\partial z_2}{\partial x_2} = 0\\
\frac{\partial^+ z_2}{\partial x_1} &=& \frac{\partial z_2}{\partial x_1} = c\phi(cx_1)\\
\label{eq_BackwardCell}
\end{eqnarray}

Knowing these derivatives, what are the derivatives of the full
network?  Let's add the superscript to each variable indicating
which unit of the network it belongs to.   Note that the outputs
of the earlier unit become the inputs of the later unit. Consider
unit 2, which gets input from units 1 and 3. If we apply
\ref{eq_RULE_ORDRD_DER} to obtain the derivative of, for example,
$z_3^2$ w.r.t. $a$, we will get, based on the topology of
connections between the units, the following result.

\begin{equation}
\frac{\partial^+ z_3^2}{\partial a} = \frac{\partial^+
z_3^2}{\partial a^2} + \frac{\partial^+ z_3^2}{\partial
z_3^3}\frac{\partial^+ z_3^3}{\partial a^3} +\frac{\partial^+
z_3^2}{\partial z_2^1}\frac{\partial^+ z_2^1}{\partial a^2}
 \label{eq_BackwardNetworkWRTa}
\end{equation}
Obviously $a^2=a^3=a$ as we use identical units. The quantities
$\frac{\partial^+ z_2^1}{\partial a^2}$ and $\frac{\partial^+
z_3^3}{\partial a^3}$ are already obtained for each unit.  Since
$z_2^1=x_3^2$ and $z_3^3=x_2^2$, the quantities $\frac{\partial^+
z_3^2}{\partial z_2^1}$ and $\frac{\partial^+ z_3^2}{\partial
z_3^3}$ are equivalent to $\frac{\partial^+ z_3^2}{\partial
x_3^2}$ and $\frac{\partial^+ z_3^2}{\partial x_2^2}$, which are
also already calculated for each individual unit.  They are the
input "deltas", or the output derivatives "propagated" through the
unit backwards. In other words, when all the quantities of each
individual unit are calculated, then the total derivatives of the
outputs of the full network w.r.t. any parameter are obtained by
summing the individual unit's derivative multiplied by the
corresponding "delta".  The correspondence is determined by the
topology of connections - knowing which output is connected to
which input. Every time we backpropagate through a unit, we also
set the values of the "deltas" of preceding units.  In our
example:

\begin{eqnarray}
\delta_2^2 &=& \frac{\partial^+ z_3^2}{\partial z_2^1} = \frac{\partial^+ z_3^2}{\partial x_3^2}\\
\delta_3^2 &=& \frac{\partial^+ z_3^2}{\partial z_3^3} = \frac{\partial^+ z_3^2}{\partial x_2^2}\\
\label{eq_DeltasInExample}
\end{eqnarray}

And

\begin{equation}
\frac{\partial^+ z_3^2}{\partial a} = \frac{\partial^+
z_3^2}{\partial a^2} + \delta_3^2 \frac{\partial^+ z_3^3}{\partial
a^3} +\delta_2^2 \frac{\partial^+ z_2^1}{\partial a^2}
 \label{eq_SumTotalDerivativeExample}
\end{equation}

Likewise, in general case, the final calculation for any network
output $j$ is a simple summation given by Eq.
\ref{eq_SumTotalDerivativeOfNetwork}.

\newpage
\section{Figure Captions}

\begin{itemize}

\item Figure \ref{fig_AlgoDerivativesInRNN}: Algorithm for
calculating ordered derivatives in recurrent neural network.  See
section \ref{section_BackProp} for explanations.

%\item Figure \ref{fig_RecurrentNetwork}: Recurrent network can be
%thought of as a feed-forward function $Net(\vec X, \vec \alpha)$
%with recurrent connections between its n outputs and m inputs. The
%recurrent connections have unit weights, effectively adding the
%output to the corresponding input.  See section
%\ref{section_BackProp} for details.

\item  Figure \ref{fig_CSRNArchitecture}: Generic Cellular SRN
architecture.

\item Figure  \ref{fig_CellGMLP}:  Cell of CSRN is a generalized
MLP with m inputs and n outputs.  The solid lines are adjustable
weights and the dashed lines are unit weights.  Note that the
output of the cell is scaled by the output weight.

\item Figure  \ref{fig_MazeWorld}:  Example of 5 by 5 maze world.
Black squares are obstacles. X is the location of the goal. The
agent needs to find the shortest path from any white square to the
goal.

\item Figure \ref{fig_MazeWithArrows}: Comparison of the solution
given by the network and the true solution.  A - approximate
solution, black arrows point in the wrong direction.  B - exact
solution.

\item Figure \ref{fig_ConnectednessExamples}: Examples of input
patterns for Connectedness problem for 7 by 7 image.

\item Figure \ref{fig_GoodExampleAB}: A. Average sum squared error
for training on 30 mazes and testing on 10. Solid - EKF training
error, dotted - EKF testing error, dashed - ALR training error,
dash-dot - ALR testing error. The 12.25 threshold for sum squared
error is shown by solid line.  B. Average Goodness of Navigation G
for training on 30 mazes and testing on 10. Solid - EKF training ,
dotted - EKF testing, dashed - ALR training, dash-dot - ALR
testing.  The 50 percent solid line is the chance level network.

\item Figure \ref{fig_CSRNAlgorithm}: Pseudo-code for the training
cycle of the CSRN.

 \item Figure \ref{fig_ExampleForAppendixA}:
Simple feedforward network (A) which can be divided into 4 blocks
(B).

\end{itemize}

\newpage
\section{Figures}

\begin{figure}[htp]
\begin{tabular}{ll}\\
\hline\\
Calculation of derivatives in recurrent network\\
\hline\\
$\delta_k^N = 1, k=1..n$ &Initialize deltas\\
$F^k_\alpha = \partial z_k^{N}/\partial \alpha, k=1..n$ &Initialize ordered derivatives\\
For t=N-1 down to 1 & for each time step\\
\hspace{15mm}For k=1 to n & for each network output\\
\hspace{20mm}  $\delta_k^t = \sum_{p \in fan-in(k)} \partial z_k^{t+1} / \partial x_p^t $& find deltas\\
\hspace{20mm}  $F^k_\alpha = F^k_\alpha + \delta_k^{t} \partial z_k^{t} / \partial \alpha $& update derivatives\\
\hline\\
\end{tabular}
\caption{} \label{fig_AlgoDerivativesInRNN}
\end{figure}

%\begin{figure}[htp]
%\centerline{\includegraphics[width=5in]{RecurrentNet.eps}}
%\caption{} \label{fig_RecurrentNetwork}
%\end{figure}

\newpage

\newpage

\newpage

%\begin{figure}[htp]
%\centerline{\includegraphics[width=5in]{KalmanBatchWeightUpdate.eps}}
%\caption{} \label{fig_KalmanBatchUpdate}
%\end{figure}
%\newpage

\newpage

\newpage

\newpage

\newpage

\begin{figure}[htp]
\begin{tabular}{l}
 \hline
Training Cycle of CSRN with EKF weight update\\
 \hline
Initialize network weights \\
Initialize EKF parameters Q, R, K\\
Repeat Until Training is Completed\\
\hspace{10mm}Set Jacobian C to empty matrix\\
\hspace{10mm}For each Training Pattern\\
\hspace{15mm}Run forward update of CSRN\\
\hspace{15mm}Calculate Network Output(s) and Error\\
\hspace{15mm}Back propagate Error though Output Transformation\\
\hspace{15mm}Backpropagate Deltas from Output Transformation through CSRN\\
\hspace{15mm}Augment the Jacobian matrix C\\
\hspace{10mm}Calculate weight adjustments using EKF algorithm\\
\hspace{10mm}For each Testing Pattern (Stopping Criteria)\\
\hspace{15mm}Run forward update CSRN\\
\hspace{15mm}Calculate Network Output\\
\hspace{10mm}Determine whether training is completed\\
\hline\\
\end{tabular}
\caption{} \label{fig_CSRNAlgorithm}
\end{figure}
\newpage

\newpage

\end{document}